\documentclass[10pt,twocolumn,letterpaper]{article}

\usepackage[pagenumbers]{wacv} 
\usepackage{graphicx}
\usepackage{amsmath}
\usepackage{amssymb}
\usepackage{bbm}
\usepackage{booktabs}
\usepackage{algorithm}
\usepackage{algpseudocode}
\usepackage[pagebackref,breaklinks,colorlinks]{hyperref}

\usepackage[capitalize]{cleveref}
\crefname{section}{Sec.}{Secs.}
\Crefname{section}{Section}{Sections}
\Crefname{table}{Table}{Tables}
\crefname{table}{Tab.}{Tabs.}

\def\wacvPaperID{417} 
\def\confName{WACV}
\def\confYear{2024}

\begin{document}
\title{HD-Fusion: Detailed Text-to-3D Generation Leveraging Multiple Noise Estimation}

\author{Jinbo Wu~~~ Xiaobo Gao~~~ Xing Liu~~~ Zhengyang Shen~~~ Chen Zhao~~~ \\
Haocheng Feng~~~ Jingtuo Liu~~~ Errui Ding \\
Department of Computer Vision Technology (VIS), Baidu Inc., China\\
{\tt\small \{wujinbo01,gaoxiaobo,liuxing12,shenzhengyang01,zhaochen03,}\\
{\tt\small fenghaocheng,liujingtuo,dingerrui\}@baidu.com}
}

\maketitle

\begin{figure*}[t]
\centering
\includegraphics[width=\textwidth]{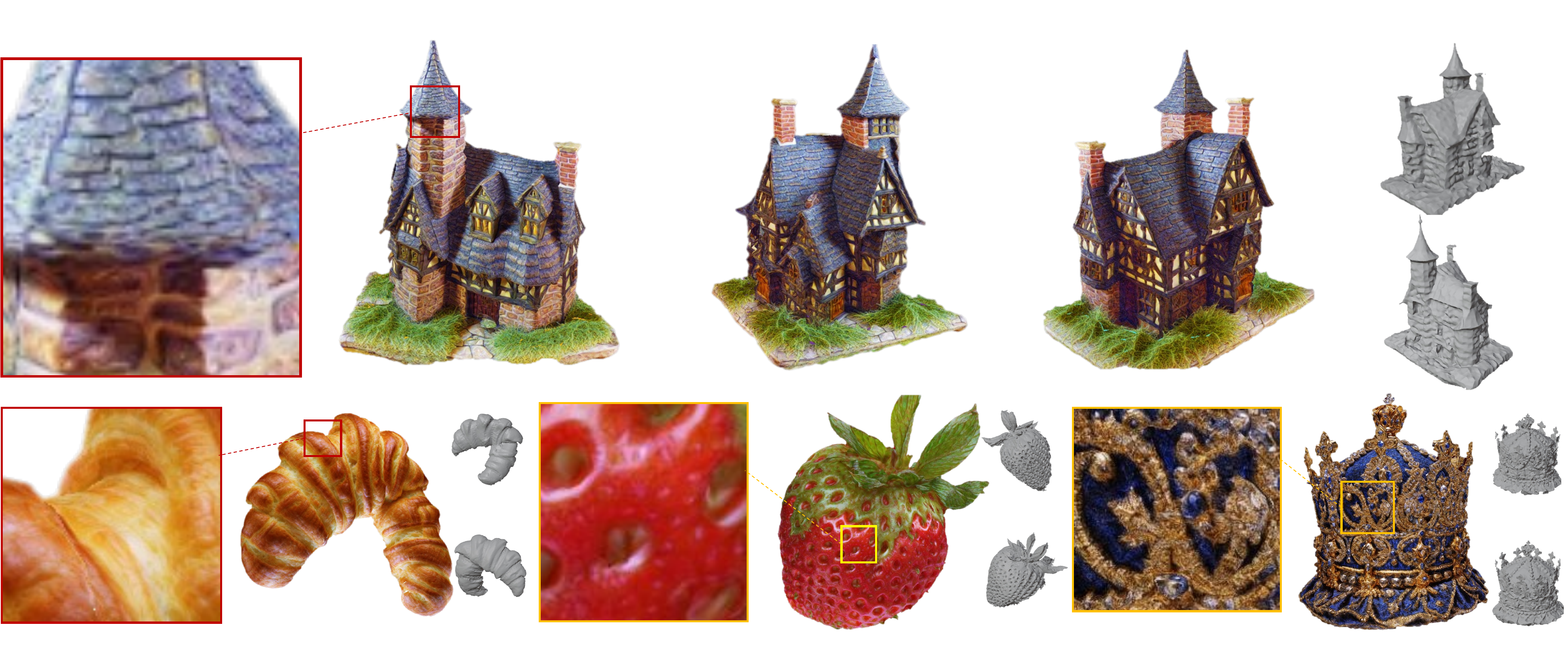}
\caption{The proposed multiple noise estimation approach yields high-quality 3D models with enhanced details. The prompts for the three contents are 1) ``A model of a house in Tudor style.'', 2) ``A delicious croissant.'' and 3) ``a ripe strawberry.''}
\label{fig:paper_face}
\end{figure*}

\begin{abstract}
\label{sec:abstract}
In this paper, we study Text-to-3D content generation leveraging 2D diffusion priors to enhance the quality and detail of the generated 3D models. Recent progress \cite{magic3d} in text-to-3D has shown that employing high-resolution (e.g., $512 \times 512$) renderings can lead to the production of high-quality 3D models 
using latent diffusion priors. To enable rendering at even higher resolutions, which has the potential to further augment the quality and detail of the models, we propose a novel approach that combines multiple noise estimation processes with a pretrained 2D diffusion prior. Distinct from the Bar-Tal et al.s' study which binds multiple denoised results \cite{multidiffusion} to generate images from texts, our approach integrates the computation of scoring distillation losses such as SDS loss and VSD loss which are essential techniques for the 3D content generation with 2D diffusion priors. We experimentally evaluated the proposed approach. The results show that the proposed approach can generate high-quality details compared to the baselines.
\end{abstract}

\section{Introduction}
\label{sec:intro}
Generating 3D content from text (Text-to-3D) is a crucial technique to support the burgeoning popularity of the Metaverse. This technology allows people to create various items by expressing their ideas through text and interacting with their creations using AR/VR headsets. Such a technique elevates the user experience to a new level of immersion.
Ideally, 3D content generated through an optimal algorithm should not only exhibit high quality in terms of geometry and appearance but also maintain a diverse range within constraints. Considering the substantial success that 2D diffusion models have achieved in text-to-image synthesis tasks, it is intuitive to conceive an approach for addressing 3D content generation that involves employing a 3D diffusion model, which would be trained and utilized in a manner similar to 2D diffusion models. However, its potential is limited by the fact that 3D diffusion-based approaches necessitate substantial computational resources and an extensive collection of 3D models paired with text for training to achieve high-quality and diverse generations. To address these difficulties, techniques for generating 3D content using 2D diffusion priors \cite{dreamfusion,latent-nerf,magic3d} have gained popularity and experienced rapid development in recent months. The advancement of this line of approaches is grounded in DreamFusion \cite{dreamfusion}, which proposed a score distillation loss, termed SDS loss. Given the images rendered from a neural radiance field $\theta$ representing a 3D object to be optimized, the SDS loss approximates the gradient directions with respect to $\theta$ for each image $I$. 
By moving $\theta$ along with the directions, the 3D object is optimized in such a way that, when rendered from a random angle, it looks as though it has been generated by the 2D diffusion model. Although DreamFusion has significantly advanced development, it falls short in generating high-quality 3D models. Lin et al. \cite{magic3d} addressed this issue, positing that the quality of 3D models can be enhanced by rendering high-resolution images to compute SDS loss. To increase training efficiency, they introduced the application of DMTet \cite{dmtet} to represent a model's shape, and proposed employing a latent diffusion model for supervision. Building on Lin et al.'s observation that high-resolution renderings contribute to improved quality in the generated 3D models, we propose an approach that allows for memory-efficient training when rendering images at even higher resolutions. In each training iteration utilizing our approach, we abstain from estimating noise for the entire rendered image, as this leads to memory issues when the image is rendered at a high resolution (e.g., $1024 \times 1024$). Instead, we crop the latent noisy image into overlapping tiles and conduct independent noise estimation processes for these tiles. Then, we integrate the estimated \textit{tiled noises} into a whole with pixel-wise weights for the computation of the SDS loss. Given that the SDS loss can be computed without the need for a computational graph of the weights in the diffusion-UNet model, we can calculate the loss without the necessity for additional GPU memory. Along with our main contribution, we present our entire Text-to-3D pipeline in this paper. We adopt a two-stage framework similar to Magic3D \cite{magic3d}'s, and incorporated ControlNet to mitigate the Janus problem and surface inconsistency. Our experiments demonstrated that 
the proposed approach improved the performance of 3D content generation. Furthermore, we highlight that the proposed method can be compatible with recent advancements such as VSD \cite{prolific-dreamer} in Text-to-3D approaches, indicating that a collaboration between these innovations and our method holds the potential for achieving more significant improvements in visual quality.
Our contribution is summarized as follows:
\begin{itemize}
\item We propose an approach that combines multiple noise estimation processes, enabling memory-efficient training for 3D generation within a high-resolution rendering space. 
\item We present an entire Text-to-3D system that leverages the proposed approach and ControlNet for geometrically correct 3D content generation through 2D diffusion priors. We provide a detailed explanation of the system.
\end{itemize}

\section{Related Work}
\label{sec:related_work}
{\bf 2D diffusion models.} Generative image models have rapidly developed \cite{glide,sde,beatgans,relate1,relate2,relate3,relate4}. Ho et al. \cite{diffusion} proposed a diffusion-based model for 2D image synthesis, which showed remarkable success in achieving high fidelity and diversity. Rombach et al. \cite{latent-diffusion} extended this technique by proposing its application in the latent space to enhance efficiency. Recently, methods for effectively fine-tuning diffusion models have garnered attention. Zhang et al. \cite{controlnet} proposed a neural network structure named ControlNet, which can be trained end-to-end with limited computational resources. The ControlNet enables a large diffusion model to be controlled by task-specific conditions (e.g., pose, normal, etc.) in addition to text. Hu et al. \cite{lora} proposed a low-rank adaptation method (LoRA) which is primarily used for fine-tuning large language models. This method was subsequently adapted for the 2D diffusion models, enabling them to be effectively fine-tuned to generate images with specific content or styles. Bar-Tal et al. \cite{multidiffusion} proposed a multi-diffusion pipeline for a controllable generation. 
Their approach binds together multiple diffusion generation processes with a shared set of parameters or constraints.
\vspace{4pt} \\
{\bf 3D representation.} Effective 3D representations have been a subject of study for many years \cite{3d-relate1,3d-relate2,3d-relate3,3d-relate4,3d-relate5,3d-relate6,3d-relate7,3d-relate8}. Recently, neural field-based approaches have demonstrated significant potential. Mildenhall et al. \cite{nerf} introduced NeRF, which implicitly represents a 3D object or a scene and generates novel views through a volume rendering algorithm given specific camera poses. In subsequent research, Muller et al. \cite{instant-ngp} proposed an advanced multi-resolution hash-encoding approach to encode the position of a 3D point. As this approach substantially accelerates the convergence during the training of a neural field, it has been widely adopted in later studies for positional encoding, including ours. Wang et al. \cite{neus} proposed an approach that represents the surface of an object using an implicit signed distance function, which is optimized using posed sparse images of the object. On the other hand, Shen et al. \cite{dmtet} proposed a deformable tetrahedral grid model to represent the geometry of a 3D object explicitly. This model can be directly optimized for accurate shape reconstruction and is made possible for mesh generation in a differentiable manner. 
\vspace{4pt}\\
{\bf Text-to-3D with 2D diffusion priors.} Poole et al. \cite{dreamfusion} proposed a loss function (SDS loss) that enables the training of 3D object generation using 2D diffusion models. Lin et al. \cite{magic3d} proposed a two-stage approach, based on the SDS loss, for generating 3D objects in a coarse-to-fine manner. They demonstrated that their approach could produce higher-quality 3D results with lower computational costs. Metzer et al. \cite{latent-nerf} introduced how they applied the computation of SDS loss in the latent space. Furthermore, they proposed the utilization of coarse shapes (e.g., a mesh with the coarse structure of a desired object) to guide the 3D generation process. Chen et al. \cite{fantasia3d} proposed an approach that disentangles geometry and appearance modeling. Through empirical evaluation, we found that this approach can recover finer geometries of a generated 3D object. Wang et al. \cite{prolific-dreamer} introduced variational score distillation that serves as an enhanced version of the SDS loss, addressing the issues of over-saturation, over-smoothing, and low diversity that sometimes occur in 3D results generated with SDS loss.

\begin{figure*}[t]
\centering
\includegraphics[width=.95\textwidth]{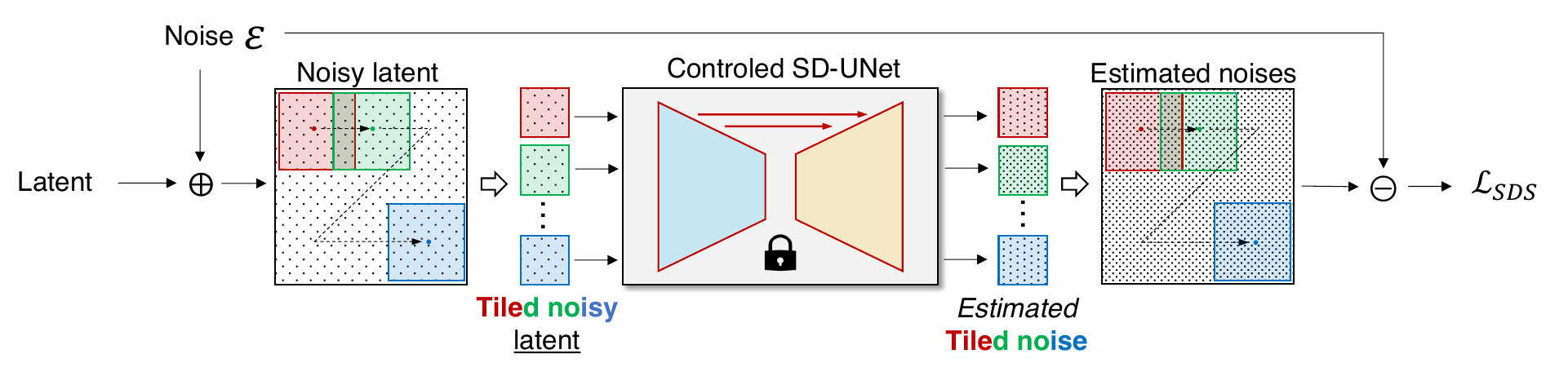}
\caption{The proposed multiple noise estimation is illustrated here. The ``Latent'' represents the latent representation of a rendered image. The ``Noise $\varepsilon$'' means the additive noise sampled at step $t$ from the diffusion process \cite{diffusion}. The ``Tiled noisy latent'' is obtained by cropping overlapping patches from the ``Noisy latent'' with a sliding window. The ``Controlled SD-UNet'' means stable diffusion model optionally powered by an instance of ControlNet. The ``Estimated noises'' is produced by consolidating all the estimated tiles of noise. }
\label{fig:overview_alg}
\end{figure*}

\section{Approach}
In this section, we introduce the proposed approach. First, we explain how we conducted the multiple noise estimation process for computing SDS loss, which enables the supervision of 3D generation. Then, we explain our two-stage Text-to-3D generation system.

\subsection{Multiple Noise Estimation for Memory-Efficient Optimization of 3D Models}
\label{sec:2d_tile_for_3d}
We illustrated the proposed multiple noise estimation approach in Fig.~\ref{fig:overview_alg}. 
Given the latent representation $J \in \mathbb{R}^{H \times W \times C}$ of a rendered image $I$, a noise $\varepsilon \sim \mathcal{N}(0, 1)$ sampled at step $t$ is applied to $J$ to calculate the noisy latent $J_{t}$ (see equ. {\color{red}4} in \cite{diffusion} ). Then, $J_{t}$ is partitioned into overlapping tiles using a square sliding window with unchanged window size, producing a set of tiled noisy latents. We employ a pretrained stable diffusion model $\phi$ (referred to as Controlled SD-UNet in Fig.~\ref{fig:overview_alg}) as our noise estimator. The lock icon means that the pretrained weights remain static during training. An instance of ControlNet is optionally employed to inject task-specific guidance into the noise estimator. The noise estimator produces the \textit{estimated noise} for each tile. We align all the estimated noises to their corresponding locations on $J_{t}$ and consolidate them into a single noise tensor with pixel-wise weights. Inspired by \cite{multidiffusion}, we set the weights to calculate pixel-wise averages within the overlapping regions.
\vspace{4pt}\\
{\bf Formulation~~} We formulate the proposed approach based on 
Lin et al.\cite{magic3d}'s extension of the SDS loss. We consider $Q_{\varphi}(\cdot~;y, t)$ as a noise estimator where $\varphi$, $y$, and $t$ denote its parameters, a condition, and a diffusion time step respectively, and consider $\mathcal{T}(\cdot)$ as an image processing function. In the proposed approach, 
we employ $\mathcal{T}(\cdot; s, k)$ as a sliding window function that outputs $M$ cropped tiles from its input with stride $s$ and window size $k$. We define $\mathcal{T}_{m}(\cdot; s, k)$ to obtain the $m^{th}$ cropped tile, which is centered at $(u_{m}, v_{m})$ with both its height and width equal to $k$. The notion $(u_{m}, v_{m})$ represents the 2D coordinates. We simplify the notion $\mathcal{T}_{m}(\cdot; s, k)$ to $\mathcal{T}_{m}(\cdot)$ for clarity.
\begin{algorithm}[!h]
\caption{The definition of the proposed multiple noise estimation function $\mathcal{M}(\cdot)$. Note, $\phi, y, t, s, k$ are omitted for clarity.}
\label{alg:noise_estimation_func}
\begin{algorithmic}[1]
\Function{$\mathcal{M}$}{$J_{t}, \xi, \mathcal{W}$} 
  \State $\xi \gets 0$ \Comment{Initialize $\xi$ to 0}
  \State $\mathcal{W} \gets 0$ \Comment{Initialize $\mathcal{W}$ to 0}
  \For{$m = 1, \ldots M$}
    \State Run $\mathcal{A}_{\xi}(m)$ and $\mathcal{A}_{\mathcal{W}}(m)$ 
    \Comment{See equ. \ref{equ:assign_xi} and \ref{equ:assign_W}}
  \EndFor
  \State  $\hat{\varepsilon} \gets \frac{\xi}{\mathcal{W}}$
   \State \Return $\hat{\varepsilon}$
\EndFunction    
\end{algorithmic}
\end{algorithm}
Consequently, we formulate the single noise estimation process on the $m^{th}$ tile as follows:
\begin{equation}
 \mathcal{F}_{\phi}(J_{t}, m)  \triangleq Q_{\phi} \circ \mathcal{T}_{m}(J_{t};y, t).
\end{equation}
We omit the notions $t$ and $y$ in $\mathcal{F}_{\phi}(\cdot)$ for the sake of clarity. To formulate the proposed \textit{multiple noise estimation}, we further define two variables which are i) $\xi \in \mathbb{R}^{H \times W \times C}$ and ii) $\mathcal{W} \in \mathbb{R}^{H \times W \times C}$, both of which are initialized by being filled with zeros. We employ $\xi_{(u_m, v_m, k)}$ and $\mathcal{W}_{(u_m, v_m, k)}$ to represent the regions centered at $(u_m, v_m)$ and enclosed by a square with its side length equal to $k$, for the two variables. Additionally, we define two assignment operators, $\mathcal{A}_{\xi}(\cdot)$ and $\mathcal{A}_{\mathcal{W}}(\cdot)$, formally expressed as:
\begin{equation}
\label{equ:assign_xi}
  \mathcal{A}_{\xi}(m) : ~ \xi_{(u_m, v_m, k)} = \mathcal{F}_{\phi}(J_{t}, m) + \mathcal{T}_{m}(\xi),
\end{equation}
\begin{equation}
\label{equ:assign_W}
  \mathcal{A}_{\mathcal{W}}(m) : ~ \mathcal{W}_{(u_m, v_m, k)} = \sum_{m = 1}^{M} \mathbbm{1} + \mathcal{T}_{m}(\mathcal{W}). \\
\end{equation}
The two operators replace the values within $\xi_{(u_m, v_m, k)}$ and $\mathcal{W}_{(u_m, v_m, k)}$ with the results computed on the right-hand side of the equalities, as per equations \ref{equ:assign_xi} and \ref{equ:assign_W}. We define the proposed multiple noise estimation function $\mathcal{M}(\cdot)$ in Alg.~\ref{alg:noise_estimation_func}, and re-formulate the SDS loss from Lin et al. \cite{magic3d}'s version as follows:
\begin{equation}
     \bigtriangledown _{\theta}\mathcal{L}_{SDS}(\phi, g(\theta)) \equiv \mathbb{E}_{t, \varepsilon} \left[
    \omega(t)(\mathcal{M}(J_{t}, \xi, \mathcal{W}) - \varepsilon) \frac{\partial{J}}{\partial{I}} \frac{\partial{I}}{\partial{\theta}}
    \right]
\end{equation}
where $I$ denotes an image rendered using $g(\cdot)$, $g(\cdot)$ represents a rendering process.
$t$ denotes a diffusion time step. $w(t)$ is a function that outputs a weight for $t$. $J$ and $J_{t}$ denote the latent image and the noisy latent image calculated at step $t$. $y$ denotes the condition(s). $\varepsilon$ denotes a noise tensor sampled according to $\mathcal{N}(0, 1)$. Through experimental validation, we demonstrate that the proposed approach effectively harnesses the strengths of high resolution in enhancing local quality and detail.

\subsection{Two-stage Text-to-3D System}
We outline the proposed two-stage Text-to-3D generation system in this section, detailed explanations for each stage are provided in subsequent sections (Sec.~\ref{sec:stage1} and Sec.~\ref{sec:stage2}). 
\vspace{4pt}\\
{\bf Outline~~} The proposed Text-to-3D system operates in two stages, aiming to generate 3D models in a coarse-to-fine manner. In the first stage, we employ a neural field to represent the shape and color of a 3D object. To encode the images generated by the neural field to latent space, we adopt the encoder of a pretrained VQ-GAN \cite{vqgan}, and compute SDS loss to optimize the neural field in latent space. In the second stage, we adopt a DMTet model and a color network to represent the 3D model. We adopt a differentiable pipeline \cite{nvdiffrast} to render the views of the model. Similar to the first stage, we encode the views using the same encoder and compute SDS loss in the latent space. In both stages, an identical diffusion model is employed for a task. We select the most appropriate diffusion model from its various versions based on performance metrics in the corresponding image generation task. To guarantee the generation of geometrically accurate views through the diffusion model, we optionally inject task-dependent guidance via ControlNet. For instance, in the case of generating 3D figures, we opt for the ``deliberate'' version of the 2D diffusion model and employ pose guidance. 
\begin{figure}[!h]
\centering
\includegraphics[width=\linewidth]{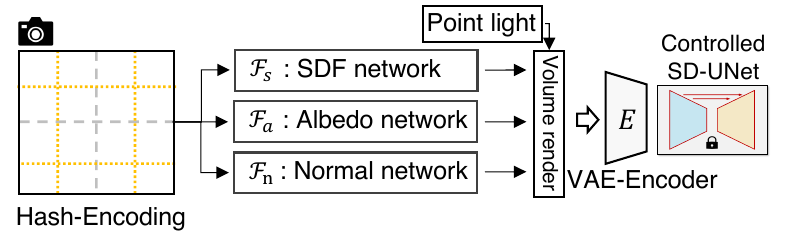}
\caption{The coarse stage of the proposed approach.}
\label{fig:stage1-pipe}
\end{figure}
\subsection{Stage1: Coarse Level Generation}
\label{sec:stage1}
In this stage, our objective is to generate a coarse representation of an object's shape and color by a neural field. We illustrate our approach in Fig.~\ref{fig:stage1-pipe}. We employ a signed distance function (SDF) $\mathcal{F}_{s}$ in conjunction with an albedo prediction network $\mathcal{F}_{a}$ and a normal prediction network $\mathcal{F}_{n}$. These modules are implemented using Multi-Layer Perceptrons (MLPs). For computational efficiency, we use one layer for each module.
To accelerate convergence, we incorporate a Hash Grid \cite{instant-ngp} for positional encoding. We set the resolution of the output images from the volume renderer to be $256 \times 256$, which we posit is an appropriate configuration for learning the coarse representation. 
\vspace{4pt}\\
{\bf Initialization~~} The modules $\mathcal{F}_{s}$, $\mathcal{F}_{a}$, and $\mathcal{F}_{n}$ are randomly initialized. The VAE-Encoder is initialized using the pretrained weights provided by \cite{latent-diffusion}. To facilitate content generation concentrated around the center of the 3D coordinate space, we initialize a ``density blob'' at the space center and utilize it as a spatial bias during training. It is noteworthy that our density blob is modeled on SDF values. This diverges from the modelings used in the previous studies \cite{dreamfusion,magic3d}, which calculated their density blobs using the densities predicted in the neural fields. We introduce how we utilize the density blob during training in the following paragraph.
\vspace{4pt}\\
{\bf Training~~} We employ a spherical coordinate system for camera placement, which is conducive to center-focused imaging. We place cameras at random locations within a limited dome and generate shaded colors for each camera to render an image. Specifically, assuming a point light source characterized by its 3D coordinate $l$, color $l_{\rho}$, and ambient light color $l_{\alpha}$, we sample 3D points along the emitting rays and calculate the color $c$ for each sampled point using diffuse reflectance \cite{lambert}:
\begin{equation}
\label{equ:lambert_render}
c = \rho \circ (l_{\rho} \circ max(0, n \cdot (l - \mu) / \left\| l - \mu \right\|) + l_{\alpha})
\end{equation} 
The albedo, denoted as $\rho$, and the normal, denoted as $n$, are predicted by the normal prediction network and the albedo prediction network, respectively. We employ a ``textureless'' rendering approach (as used in \cite{dreamfusion}) and incorporate it at a specific ratio to enhance the robustness of the training process. We calculate the aforementioned spatial bias using
\begin{equation}
\tau_{init}(\mu) = \lambda_{\tau} \cdot \left(||\mu - r ||^{2}\right),
\end{equation}
where $\mu$ is the norm of a point sampled along a ray, $r$ denotes the radius of the ``density blob'', and $\lambda_{\tau}$ denotes the scale parameter. $\tau_{init}(\mu)$ is the calculated bias added to the predicted SDF value of the sampled point. We set $\lambda_{\tau} = 1.0$ and $r = 0$ in our experiment. We omit the introduction to volume rendering techniques as they are already detailed in previous studies such as \cite{neus}.
We do not use the proposed multiple noise estimation processes during coarse-level training, as we do not expect high-quality details to be learned at this stage. 

\subsection{Stage2: Fine Level Generation}
\label{sec:stage2}

\begin{figure}[h]
\centering
\includegraphics[width=\linewidth]{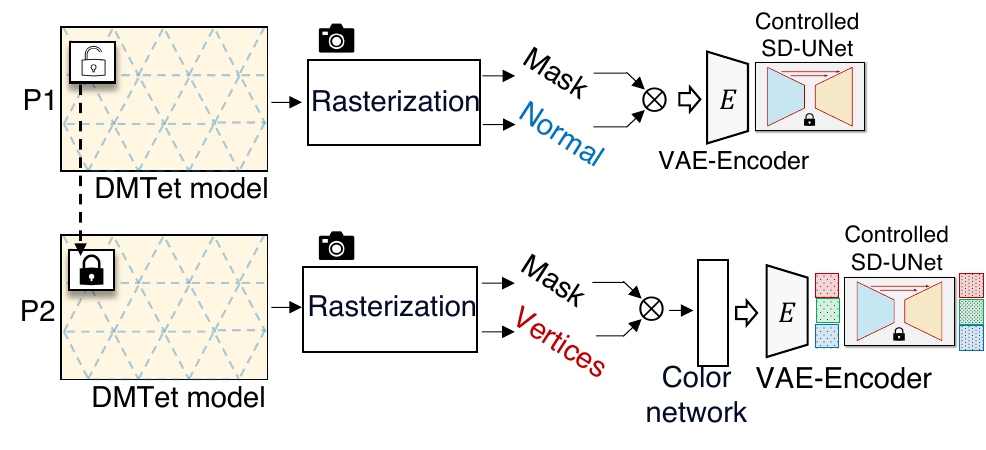}
\caption{This figure illustrates the fine-level generation stage of the proposed approach. It has two phases, denoted by P1 and P2. We learn geometry and color in P1 and P2 separately. The DMTet model is optimized in P1 and is fixed in P2. The proposed multiple noise estimation is only applied in P2.}
\label{fig:stage2-pipe}
\end{figure}

We present how we generate high-quality 3D models based on their coarse shapes and colors. The pipeline of this stage is illustrated in Fig.~\ref{fig:stage2-pipe}. We utilize a Deformable Marching Tetrahedral Grid (DMTet) model \cite{dmtet} to represent an object's geometry due to its efficiency. To represent the colors, we employ a color network that is implemented in the same manner as the albedo network in Stage 1. We employ a rasterization model to output the normal and the indices corresponding to the set of vertices in the DMTet model. The VAE-Encoder takes the rendered image as input and outputs its latent representation. We then compute the SDS loss within the latent space using the diffusion model (depicted as Controlled SD-UNet in Fig.~\ref{fig:stage2-pipe}). It should be noted that we learn geometry (depicted as P1 in Fig.~\ref{fig:stage2-pipe}) and colors (depicted as P2 in Fig.~\ref{fig:stage2-pipe}) separately. The proposed multiple noise estimation is employed only in P2. We explain the details in the following paragraph.
\begin{figure*}[t]
\centering
\includegraphics[width=.70\textwidth]{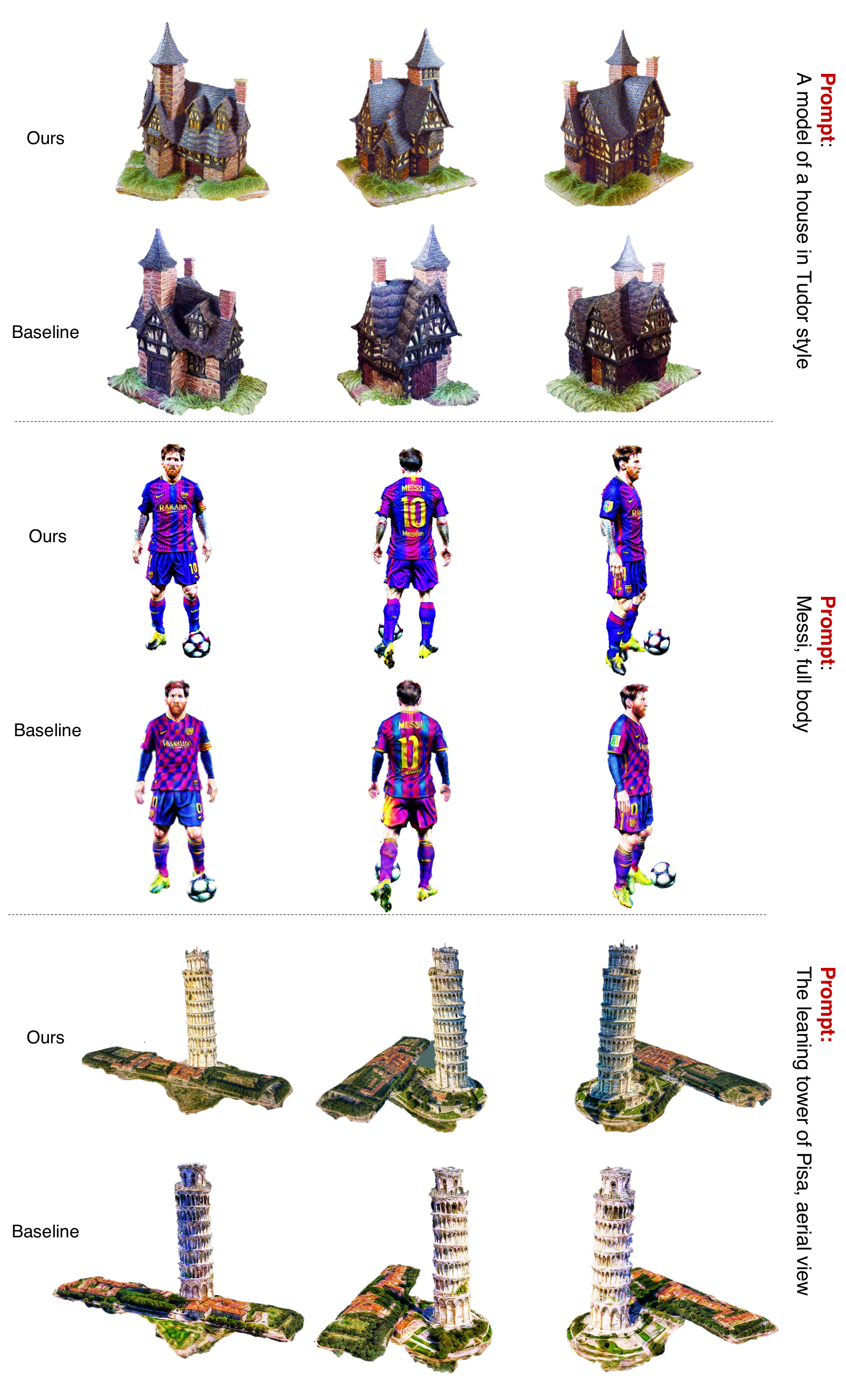}
\caption{Result comparison. ``Ours'' means the results generated {\bf with} the proposed multiple noise estimation. ``Baseline'' means the results generated \underline{without} applying the multiple noise estimation. }
\label{fig:compre_w_baseline}
\end{figure*}

\begin{figure*}[t]
\centering
\includegraphics[width=.95\textwidth]{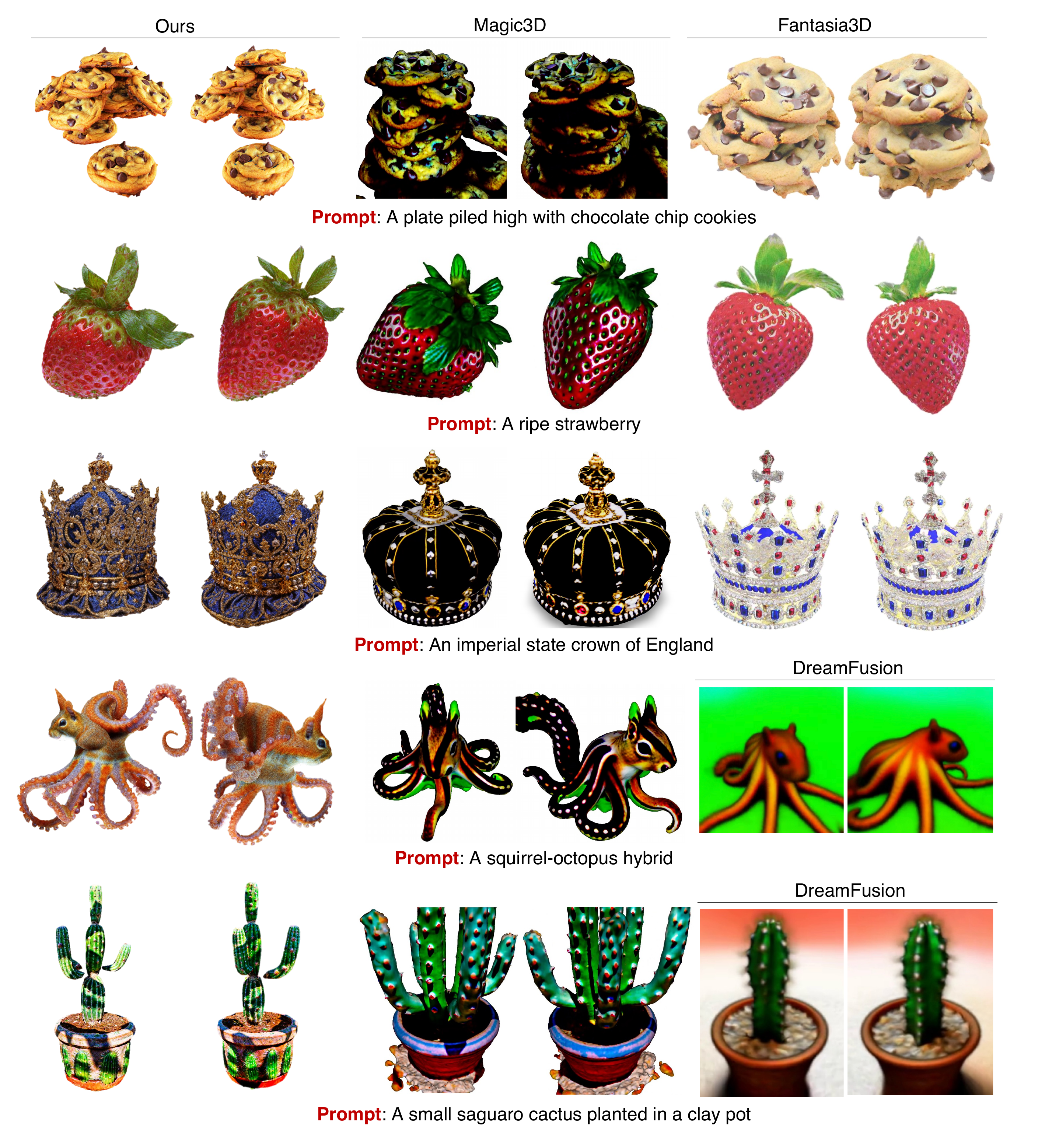}
\caption{Result comparison. ``Ours'' means the results generated by the proposed approach. Matic3D \cite{magic3d}, Fantasia \cite{fantasia3d} and DreamFusion \cite{dreamfusion} are the SOTA methods. }
\label{fig:compre_for_sota2}
\end{figure*}


\subsubsection{Initialization}
{\bf DMTet model~~} To initialize the DMTet model, we set the SDF values of the vertices by querying the neural SDF field, which has been trained in Stage 1. We assign zeros to the  offsets for all the vertices. This process allows the DMTet model roughly captures the shape of the object. 
\vspace{4pt}\\
{\bf Color network~~} We initialize the color network by employing the weights obtained from the color network that was trained in Stage 1.
\\
\subsubsection{Training}
We first train our model to learn geometry (denoted by ``P1'' in Fig.~\ref{fig:stage2-pipe}) until it converges, then we freeze the geometry and train it to learn colors (denoted by ``P2'' in Fig.~\ref{fig:stage2-pipe}). We employ the spherical coordinate system to position our cameras, enabling them to focus on the center. 
\vspace{4pt}\\
{\bf Phase1: Learning geometry~~} Inspired by a recent advance \cite{fantasia3d}, We optimize the DMTet model using normal images rendered from various viewpoints. The rasterization module \cite{nvdiffrast} calculates the normal image and the object mask from the DMTet model given a specific camera pose. Subsequently, the mask is employed to filter out the background pixels in the normal image. The filtered normal image is then fed into the VAE-Encoder $E$ for the computation of the SDS loss in the latent space. We set the resolution as $512 \times 512$ for rendering the normal image.
\vspace{4pt}\\
{\bf Phase2: Learning Color~~} Once the geometry learning is completed in phase 1, we freeze the DMTet model and exclusively learn colors using the pipeline depicted as P2 in Fig.~\ref{fig:stage2-pipe}. To render an image, we first collect a set of corresponding vertices by utilizing the rasterization module and the object mask. Then, we query the color network using the 3D positions of these vertices. Since the rasterization process has already established the projections from the 3D coordinates to their corresponding 2D positions on the output image, the rendered image can be obtained by mapping the colors according to these projections. The VAE-Encoder encodes the rendered images into latent space to compute SDS loss. To facilitate convergence, we first train the model using the standard SDS loss with the diffusion prior, then we activate the proposed multiple noise estimation for continued training until completion
\begin{figure*}[t]
\vspace{-2cm}
\centering
\includegraphics[width=.85\textwidth]{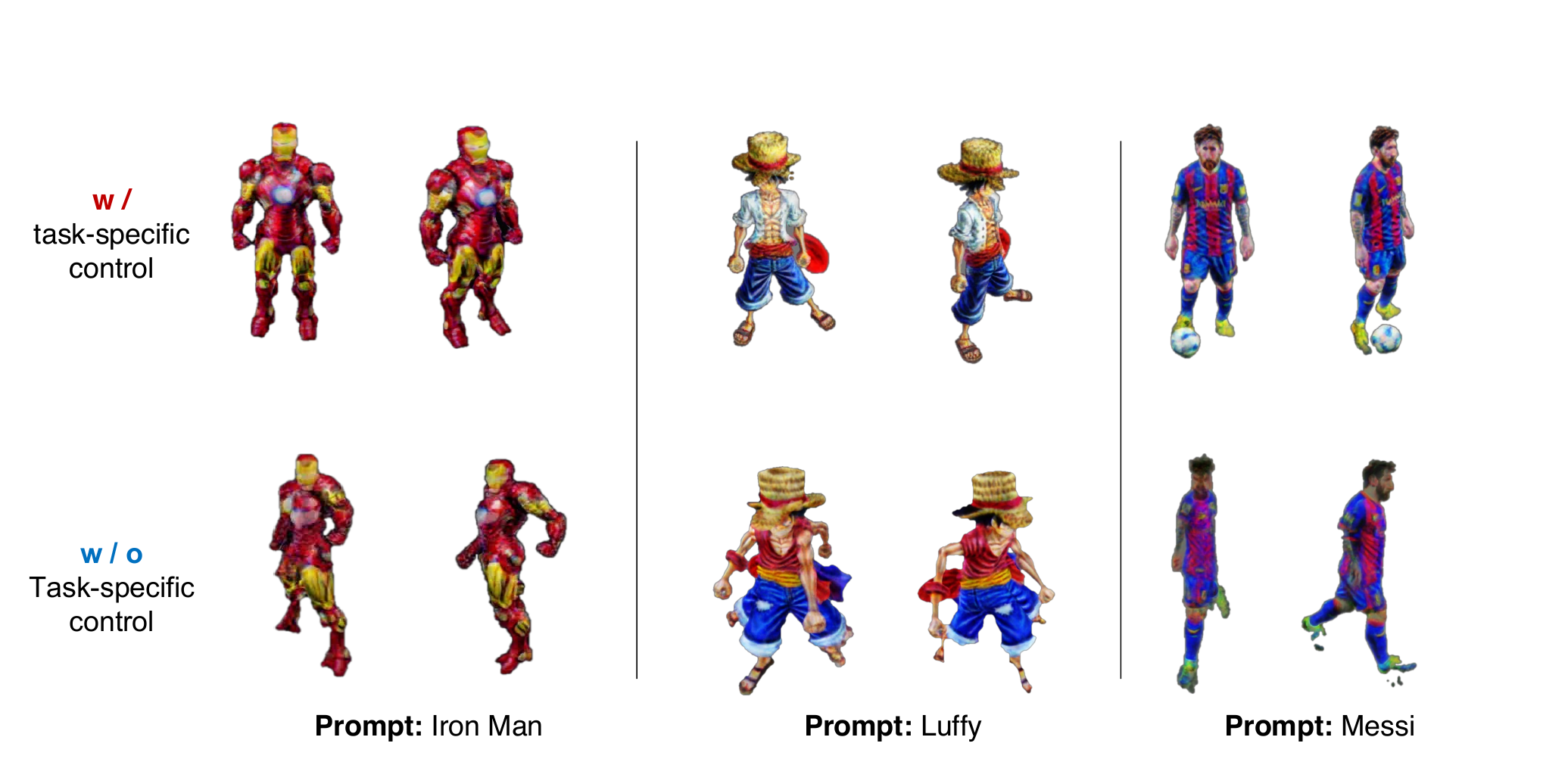}
\caption{Pose-guidance is employed using ControlNet for 3D human character generation. The results are produced from stage 1 of the proposed system. It is seen that the results (top) generated with the guidance are better than the ones (bottom) generated without the guidance. }
\label{fig:ablation_control}
\end{figure*}

\section{Experiments}
{\bf Result comparison~~} In this section, we conduct thorough experiments to evaluate the proposed approach. We first compare the results produced with and without the proposed multiple noise estimation. It is seen in Fig.~\ref{fig:compre_w_baseline} that the proposed multiple noise estimation approach outperformed the baseline in terms of high quality with enhanced details. Next, we compare the proposed approach with the SOTA methods which are Magic3D \cite{magic3d} and Fantasia3D \cite{fantasia3d}, using identical prompts. For a fair comparison with the SOTA methods,
images borrowed from their original papers are used for comparison, with our best efforts made to preserve image quality. It is seen from Fig.~\ref{fig:compre_for_sota2} that the proposed approach generates better 3D models than the SOTA methods in terms of high quality. 
\vspace{4pt}\\
{\bf Task-specific guidance~~} We optionally employ task-specific guidance in the proposed approach. We take the 3D human character generation task as an example to study how it affects performance. In the experiment, we employed pose guidance, implemented using ControlNet, to address the Janus problem. The results obtained from stage 1 of our approach are shown in Fig.~\ref{fig:ablation_control}. Notably, we omitted the training in stage 2 because the results produced from stage 1 are sufficiently valid for this study. It is observed that the inclusion of pose guidance effectively mitigates geometric distortions in the generated 3D characters. We believe task-specific guidance plays an important role in generating high-quality 3D content and is a worthwhile subject for in-depth future studies. \vspace{4pt}
\\
\begin{figure}[t]
\centering
\includegraphics[width=.85\linewidth]{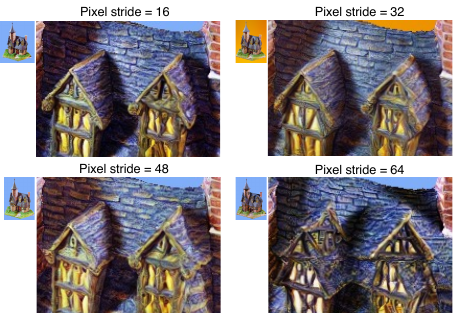}
\caption{Examples produced with four configurations of pixel stride.}
\label{fig:ablation_pixel_stride}
\vspace{-8pt}
\end{figure}
{\bf Pixel stride~~} As depicted in Fig.~\ref{fig:overview_alg}, we utilize a sliding window operator in our proposed approach to produce overlapping patches from a noisy latent. It is worthwhile to investigate how the pixel stride of the sliding window impacts the generation of high-quality details. We study this by using four distinct pixel strides: 16, 32, 48, and 64. Notably, configuring the pixel stride to 16, 32, or 48 yields overlapping tiles, while a pixel stride of 64 results in non-overlapping tiles. The results are shown in Fig~.\ref{fig:ablation_pixel_stride}. 
We observe that the results generated with overlapping tiles (that is, when the pixel stride is set to 16, 32, or 48) are comparably good. On the other hand, the result produced without overlapping tiles is of lesser quality.


\subsection{Conclusion}
In this paper, we have proposed a multiple noise estimation approach that enables memory-efficient training for 3D generation within a high-resolution rendering space. We have evaluated the proposed approach through experiments and demonstrated that the proposed approach is effective in generating high-quality 3D models with enhanced details. In addition, we have presented an entire Text-to-3D system that leverages the proposed approach and ControlNet for geometrically correct 3D content generation through 2D diffusion priors. 

{\small
\bibliographystyle{ieee_fullname}
\bibliography{egbib}
}

\end{document}